\documentclass[10pt,twocolumn,letterpaper]{article}

\usepackage[pagenumbers]{cvpr} % To force page numbers, e.g. for an arXiv version

\usepackage[dvipsnames]{xcolor}

\usepackage{graphicx}
\usepackage{amsmath,bm}
\usepackage{amssymb}
\usepackage{booktabs}
\usepackage{tikz,pgfplots}
\usepackage{color, colortbl}
\usepackage{float,array,multirow}
\usepackage[percent]{overpic} % in preamble
	
\definecolor{Gray}{gray}{0.9}
\usepackage{arydshln}
\usepackage[table]{xcolor}

\definecolor{cvprblue}{rgb}{0.21,0.49,0.74}
\definecolor{bluish_green}{rgb}{0.0,0.62,0.45}
\definecolor{sky_blue}{rgb}{0.34, 0.71, 0.91}
\usepackage[pagebackref,breaklinks,colorlinks,citecolor=cvprblue]{hyperref}

\def\paperID{315} % *** Enter the Paper ID here
\def\confName{3DV\xspace}
\def\confYear{2026\xspace}

\title{Improving 3D Foot Motion Reconstruction in Markerless Monocular\\Human Motion Capture}

\author{Tom Wehrbein \qquad Bodo Rosenhahn \\
L3S - Leibniz University Hannover, Germany\\
{\tt\small wehrbein@tnt.uni-hannover.de}\\
}

\begin{document}

\twocolumn[{%
	\renewcommand\twocolumn[1][]{#1}
    \vspace{-1.5em}
	\maketitle
	\begin{center}
		\newcommand{\teaserwidth}{\textwidth}
		\centerline{
			\includegraphics[width=\teaserwidth,clip]{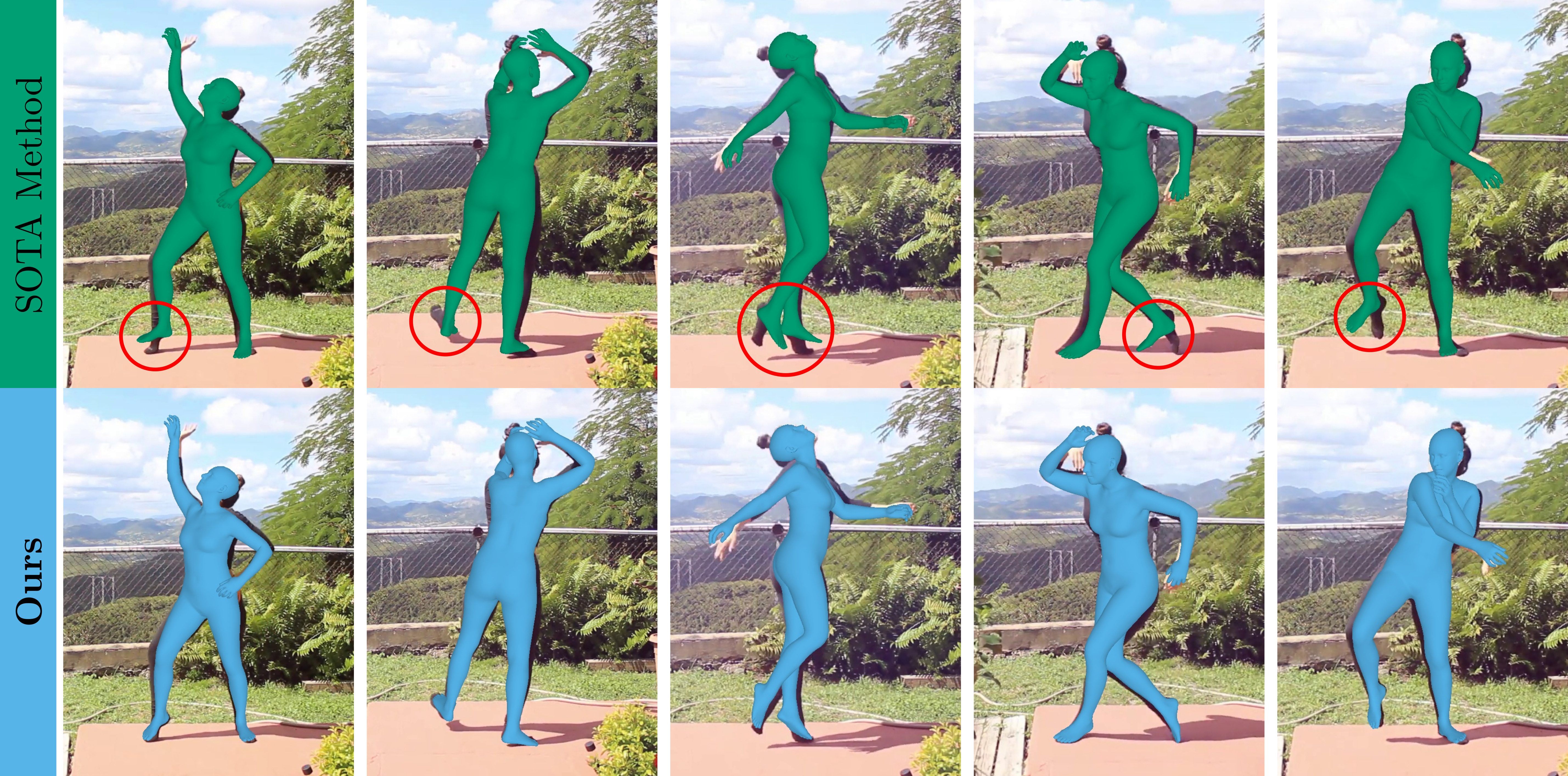}
		}
		\captionof{figure}{State-of-the-art 3D human motion recovery methods like \textcolor{bluish_green}{GVHMR}~\cite{shen2024gvhmr} \textit{(top)} fail to capture complex 3D foot movement when given in-the-wild videos.
        We identify this to be mainly an issue of inaccurate and insufficient video training data.
        To address this, we introduce {FootMR}, a Foot Motion Refinement method that leverages large-scale motion capture data to learn lifting 2D foot keypoint sequences to 3D.
        By effectively resolving ambiguities in 2D-to-3D mapping, \textcolor{sky_blue}{FootMR} \textit{(bottom)}, when combined with an existing 3D~human recovery model, generates realistic and accurate 3D foot motion, significantly outperforming previous work.
        }
		\label{fig:teaser}
        \vspace{-0.55em}
	\end{center}
}]

\begin{abstract}
State-of-the-art methods can recover accurate overall 3D human body motion from in-the-wild videos.
However, they often fail to capture fine-grained articulations, especially in the feet, which are critical for applications such as gait analysis and animation.
This limitation results from training datasets with inaccurate foot annotations and limited foot motion diversity.
We address this gap with FootMR, a Foot Motion Refinement method that refines foot motion estimated by an existing human recovery model through lifting 2D foot keypoint sequences to 3D.
By avoiding direct image input, FootMR circumvents inaccurate image–3D annotation pairs and can instead leverage large-scale motion capture data.
To resolve ambiguities of 2D-to-3D lifting, FootMR incorporates knee and foot motion as context and predicts only residual foot motion.
Generalization to extreme foot poses is further improved by representing joints in global rather than parent-relative rotations and applying extensive data augmentation.
To support evaluation of foot motion reconstruction, we introduce MOOF, a 2D dataset of complex foot movements. 
Experiments on MOOF, MOYO, and RICH show that FootMR outperforms state-of-the-art methods, reducing ankle joint angle error on MOYO by up to 30\% over the best video-based approach.
Our code and dataset are available for research purposes at \href{https://twehrbein.github.io/footmr-website/}{twehrbein.github.io/footmr-website/}.

\end{abstract}
%-------------------------------------------------------------------------------------------------------
\section{Introduction}
\label{sec:intro}
Accurately reconstructing the 3D motion of a person from monocular video has been a major research problem for several decades.
The field has advanced substantially in recent years, and current methods can robustly recover promising human motion from casual videos.
However, while the coarse body movement is usually reconstructed with high accuracy, they fail at reconstructing the fine-grained movement of the feet (see Fig.~\ref{fig:teaser}).
We argue that accurately reconstructing nuanced foot motion is particularly important for many applications in sports, medicine, AR/VR, and animation.
For example, when capturing the performance of a dancer, the movement of the feet plays a critical role in producing detailed and lifelike animation.

\begin{figure}
\centering
\includegraphics[width=1.0\linewidth]{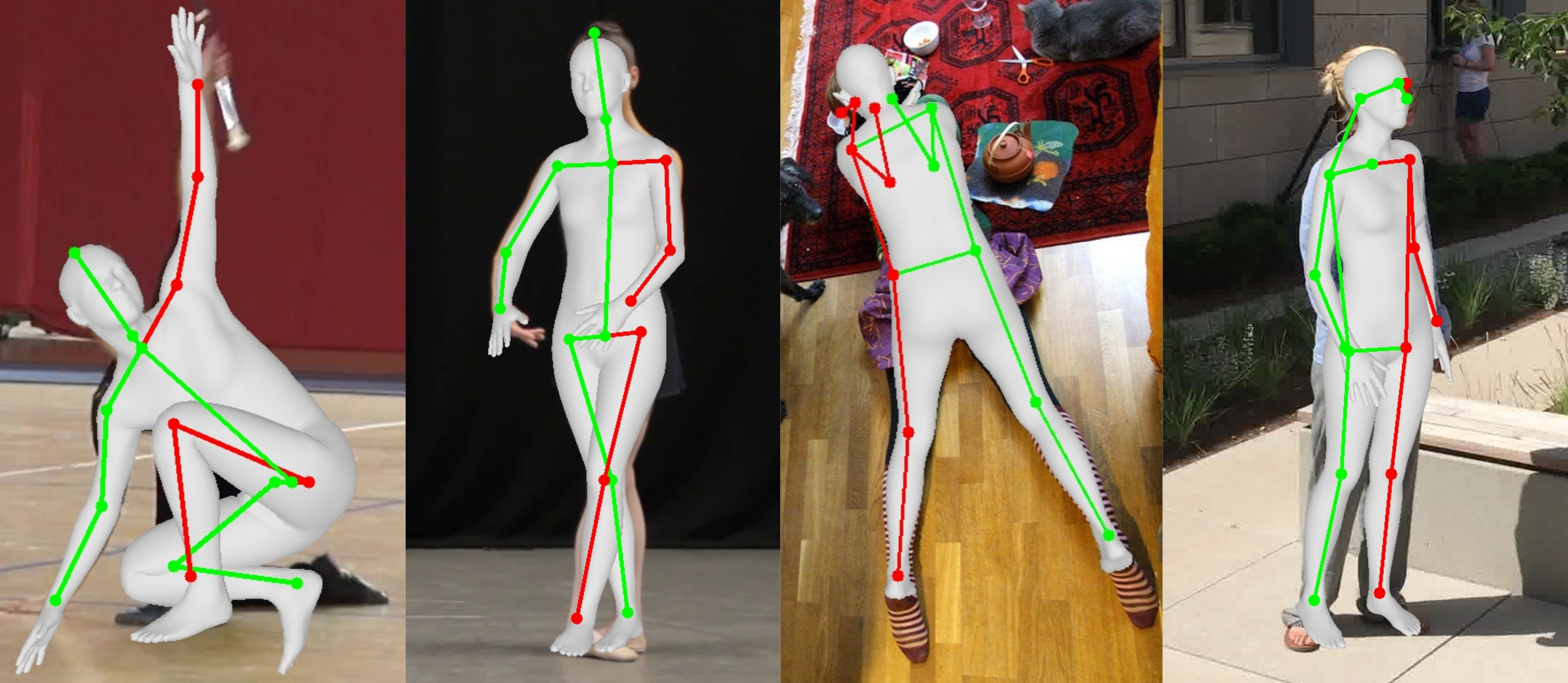}
\caption{\textbf{Erroneous 3D foot annotations} in pseudo-GT fits generated by fitting 3D models to sparse keypoints.
Images are from MPII~\cite{andriluka2014mpii}, COCO~\cite{lin2014coco}, and 3DPW~\cite{marcard2018eccv}.
Please zoom in for details.}
\label{fig:bad_gt_feet}
\vspace{-1.0em}
\end{figure}

Our insight is that previous approaches fail because they rely on in-the-wild training data with inaccurate 3D foot annotations.
Such pseudo-ground truth (pseudo-GT) annotations are generated by fitting a parametric body model to sparse 2D keypoints~\cite{goel2023humans,joo2020eft,li2022cliff,kolotouros19spin,moon2022annot}, sometimes combined with IMU data~\cite{marcard2018eccv}.
Although these fitting targets can constrain the coarse 3D body pose, details of the feet are often lost because keypoints are typically defined only for major body joints, extending no further than the ankle.
With only a single keypoint for the ankle, the 3D pose of the foot is not sufficiently constrained, leading to inaccurate pseudo-GT fits as shown in Fig.~\ref{fig:bad_gt_feet}.
However, training on large-scale, diverse data is crucial for models to generalize well to different motions and scenes.
Another challenge is that existing 3D human video datasets~\cite{marcard2018eccv,ionescu2014h36m,metha3dv3dhp,huang2022rich} mainly contain subjects performing everyday activities with very little foot motion or synthetic humans without shoes~\cite{black2023bedlam,varol17_surreal}.
This limits the ability of models to generalize to complex foot movements typical of dance, ballet, and sports.

We address these challenges with \textit{FootMR}, a \textit{Foot Motion Refinement} method that refines foot motion estimated by an existing 3D human motion recovery model.
Instead of directly using images as input, FootMR processes 2D foot keypoints from an off-the-shelf detector~\cite{khirodkar2024sapiens}.
We use four keypoints per foot: big toe, small toe, heel, and ankle.
During training, we synthetically generate 2D keypoints and ground truth 3D motion sequence pairs using both large-scale motion capture and video datasets. 
By avoiding direct image input, FootMR completely bypasses the dependency on inaccurate image-3D foot annotation pairs.
Foot articulation in parametric 3D human body models is primarily defined by rotations of the ankle joint.
Therefore, we train a model to learn lifting 2D foot keypoint sequences to 3D~ankle rotations.
Because 2D-to-3D lifting is inherently ambiguous and degenerates when input keypoints are noisy, we incorporate knee and initial ankle rotations estimated by an existing 3D human recovery model as additional input and predict only residual ankle rotations.
The knee is the parent joint of the ankle along the kinematic chain and thus provides information about the space of feasible ankle rotations.
Together with leveraging initial ankle estimates and the temporal motion context, this helps resolving ambiguity of the 2D-to-3D mapping.
To further improve generalization to extreme foot poses, FootMR processes global rather than parent-relative joint rotations.
Intuitively, because foot motion diversity in training datasets is limited, the range of possible 3D ankle rotations observed during training is very narrow when using parent-relative rotations.
Additionally, we use heavy data augmentation by applying a random 3D rotation to the root orientation of all 3D poses of a sequence.
This is possible because no images are used for training FootMR, and the 2D keypoints can be efficiently synthesized through projection with a virtual camera.
Without using additional training data or having to meticulously fix pseudo-GT annotations, FootMR accurately captures intricate 3D motion of the feet (see Fig.~\ref{fig:teaser}).
The refined ankle predictions are fused with the remaining body parameters estimated by a 3D human motion recovery model.

To support evaluation of foot motion reconstruction, we collect a new dataset with complex \textbf{MO}vements \textbf{O}f the \textbf{F}eet (MOOF).
We record videos of individuals performing simple body movements with complex foot movements, \eg, a person sitting on a chair doing ankle circles, and extend these recordings with dance and ballet videos collected online.
In total, MOOF consists of 41 videos with annotated 2D foot keypoints.
Experiments on MOOF and on the 3D datasets MOYO and RICH demonstrate that FootMR achieves significantly more accurate foot motion reconstruction than all competitors.

Our main contributions are summarized as follows:
\begin{itemize}
    \item We propose FootMR, a method that leverages 2D foot keypoints to refine 3D foot motion estimated by a human recovery model.
    \item We show that 2D-to-3D foot motion lifting works robustly when using knee and initial foot motion as context.
    \item We collect MOOF, a new video dataset containing complex foot movements with annotated 2D foot keypoints.
    \item FootMR generalizes to extreme foot poses and outperforms all competitors on MOYO, RICH, and MOOF.
\end{itemize}

%-------------------------------------------------------------------------------------------------------
\section{Related Work}
\label{sec:related_work}

\subsection{Monocular Human Mesh Recovery}
Reconstructing 3D human pose and shape from monocular images is most widely formulated as estimating the low-dimensional parameters of a statistical body model~\cite{loper2015smpl,pavlakos2019smplx,osman2022supr,joo2018totalcapture,xu2020ghum}.
Pioneering work~\cite{bogo16eccv,guan09iccv,hasler10cvpr,sigal07nips} investigates optimization-based approaches by fitting the body parameters to image observations.
Starting with HMR~\cite{kanazawa18hmr}, direct regression methods based on deep learning became the leading paradigm.
Many methods follow HMR in using a backbone to extract image features followed by a multilayer perceptron (MLP) that regresses body parameters \cite{wehrbein25humr,heo25deforhmr,li2022cliff,patel2024camerahmr,zhang21pymaf}.
Improved backbones~\cite{sun2019hrnet,dosovitskiy2021an,goel2023humans} and better camera modeling~\cite{kissos2020weakpersp,li2022cliff,kocabas21spec,patel2024camerahmr,wang2023zolly,wang2025blade} played an important role in advancing reconstruction accuracy in recent years.
As for all data-driven approaches, another key factor for accuracy and robustness is the availability of high-quality, large-scale datasets.
Since such 3D human datasets, especially in the wild, are difficult to obtain, previous work focuses on generating 3D pseudo-GT annotations.
This is done by fitting body parameters to 2D keypoints~\cite{kolotouros19spin,patel2024camerahmr,goel2023humans,wehrbein2023iccvw} using SMPLify~\cite{bogo16eccv}, or by fine-tuning a pretrained body regressor on the target images using 2D keypoints as weak supervision~\cite{li2022cliff,joo2020eft,moon2022annot}.
Although training with 3D pseudo-GT annotations is crucial for models to generalize well, we notice that they are often inaccurate for the feet (see Fig.~\ref{fig:bad_gt_feet}) and thus lead to models with good body but poor foot pose reconstructions.
CameraHMR~\cite{patel2024camerahmr} improves upon this by using a newly introduced dense surface keypoint detector to estimate more keypoints for pseudo-GT fitting.
However, it fails to generalize to extreme foot poses not seen during training.
Additionally, image-based models applied to frames of a video sequence often produce temporally inconsistent body poses and shapes.

Video-based approaches encode temporal information by jointly processing static features extracted from each frame.
Earlier methods use convolutional~\cite{kanazawa2019hmmr} or recurrent encoders~\cite{kocabas2019vibe,luo2020meva,choi2020beyond}, while transformer architectures are employed by more recent methods~\cite{wan2021maed,shen2023global,WeiLin2022mpsnet}.
Even more recently, several approaches~\cite{sun2023trace,yuan2022glamr,ye2023slahmr,li2024coin,yin2024whac,shin2024wham,wang2024tram,shen2024gvhmr,pace2024kocabas} aim to recover global human motion to handle arbitrary moving cameras.
While producing temporally coherent results and accurate body estimates, all methods fail to capture complex 3D foot movements because, similar to image-based methods, they rely on training data with inaccurate foot annotations.
Additionally, existing 3D human video datasets are much more limited in diversity than image datasets, typically containing very little foot motion.

\subsection{Foot Pose Estimation}
Most existing human pose datasets provide only minimal foot annotations, limited to the position of the ankles.
The first 2D foot keypoint dataset was released by Cao \etal~\cite{cao2019openpose} which extends a subset of COCO~\cite{lin2014coco} by labeling three keypoints per foot: big toe, small toe, and heel. 
Similarly, Jin \etal~\cite{jin2020whole} introduced COCO-WholeBody for 2D human whole-body pose estimation.
Enabled by these datasets, recent 2D detectors~\cite{khirodkar2024sapiens,xu2022zoomnas,xu2022vitpose+,jian2023rtmpose,yang2023effective} can accurately and robustly detect foot keypoints in addition to body keypoints.

To reconstruct foot keypoints in 3D, previous work~\cite{joo2018totalcapture,zhu2023h36mwhole,zhuo2023foot} relies on multi-view optimization using corresponding 2D detections.
Zhu \etal~\cite{zhu2023h36mwhole} provide their reconstructions for two existing multi-view datasets~\cite{ionescu2014h36m,joo2018totalcapture}.
Zhuo \etal~\cite{zhuo2023foot} construct annotations for training a human mesh recovery model, but do not release the annotations or the model.
Relying on multi-view indoor data is very restricting and leads to a lack of diversity in scenes and actors, limiting robustness of data-driven methods.
In contrast, by refining foot motion decoupled from the rest of the body, FootMR only requires 2D foot keypoints and can thus leverage large-scale motion capture data together with in-the-wild video datasets without suffering from inaccurate foot annotations.
During inference, FootMR benefits from the accuracy and robustness of recent 2D whole-body keypoint detectors.

Several recent works~\cite{rempe2021humor,shimada2020physcap,rempe2020contact,yi2021transpose,zou2020footskate,yi2022pip} focus on the physical plausibility of 3D human motion to mitigate artifacts such as foot sliding or foot-floor penetration.
However, they do not address the accuracy or evaluation of foot motion reconstruction, which is the focus of this work.

%-------------------------------------------------------------------------------------------------------
\section{Method}
\label{sec:method}

\begin{figure*}
\centering
\includegraphics[width=1.0\linewidth]{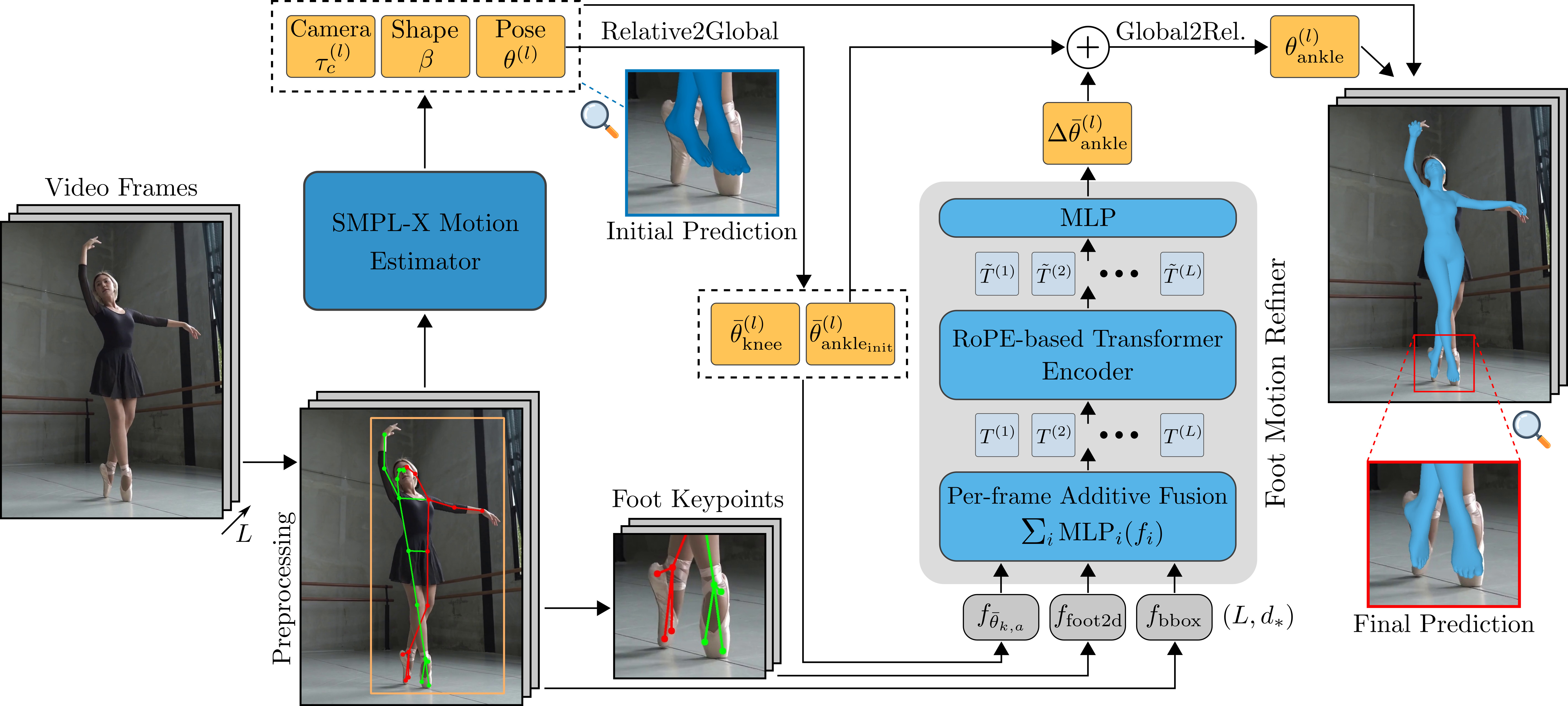}
\caption{\textbf{Overview of our framework.}
Given a monocular video, a SMPL-X motion estimator is employed to estimate 3D human motion including erroneous initial ankle rotations.
Knee and ankle predictions are then transformed from parent-relative to global rotations and used together with 2D foot keypoints and bounding boxes as input for Foot Motion Refinement (FootMR).
FootMR refines the initial ankle predictions by estimating residual rotations.
After fusing the refined ankle rotations with the remaining body parameters, the final output of our framework is accurate and temporally coherent 3D human body \textit{and} foot motion.
}
\label{fig:method}
\vspace{-1.0em}
\end{figure*}

Given a monocular video $\{I^{(l)}\}^L_{l=1}$ of length $L$, our objective is to recover 3D human motion that is accurate not only for the body but also for the feet.
We adopt SMPL-X~\cite{pavlakos2019smplx} to represent the 3D human body, which consists of $J$ joint rotations $\{\theta^{(l)} \in \mathbb{R}^{J\times6}\}^L_{l=1}$, shape parameters $\beta \in \mathbb{R}^{10}$, and translation $\{\tau_c^{(l)} \in \mathbb{R}^3\}^L_{l=1}$ in the camera space.
We use the continuous 6D rotation representation proposed by~\cite{zhou19rot6d} to represent 3D joint rotations.
To parameterize foot articulation, the SMPL-X kinematic tree contains a joint for the ankle and forefoot.
Since the ankle joint has the most influence on foot motion, and previous work fails to accurately capture it, we focus on improving motion reconstruction of the left and right ankles $\{{\theta}_{\text{ankle}}^{(l)} \in \mathbb{R}^{2\times6}\}^L_{l=1}$.

To address the challenge of insufficient video data with accurate 3D foot annotations, we decouple the reconstruction process.
First, a SMPL-X motion estimator is used to estimate 3D human motion including potentially erroneous initial ankle rotations.
The initial ankle predictions are then refined by our proposed Foot Motion Refinement (FootMR) method, which utilizes 2D foot keypoints instead of raw images and thus avoids relying on inaccurate image-3D foot annotation pairs during training.
An overview of our framework is shown in Fig.~\ref{fig:method}.

\subsection{Foot Motion Refinement}
\label{sec:foot_transformer}
\noindent\textbf{Input and preprocessing.} We design a transformer-based model to jointly refine initial left and right ankle rotations $\bar{\theta}_{\text{ankle}_\text{init}}^{(l)}\in\mathbb{R}^{2\times6}$ by predicting residual rotations \mbox{$\Delta\bar{\theta}_{\text{ankle}}^{(l)}\in\mathbb{R}^{2\times6}$} from three types of input.
The first is the 2D foot keypoint sequence $f_{\text{foot2d}} \in \mathbb{R}^{L\times16}$ that consists of four keypoints per foot: big toe, small toe, heel, and ankle.
We normalize the keypoints using the person's bounding box and provide the center and scale of the bounding boxes as second input $f_{\text{bbox}} \in \mathbb{R}^{L\times3}$~\cite{li2022cliff}.
This provides information about the location of the person in the original image which is important due to perspective effects.
Keypoints that are not visible based on their confidence score are set to zero.
The third input of FootMR is the global rotations of the left and right knee and ankle $f_{\bar{\theta}_{k,a}} \in \mathbb{R}^{L\times24}$ predicted by a SMPL-X motion estimator.
In SMPL-X, each joint rotation is defined relative to its parent joint in the kinematic tree.
Let $\theta_i^{(l)} \in \mathbb{R}^{6}$ denote the parent-relative rotation of joint $i$ in 6D representation.
Following Zhou \etal~\cite{zhou19rot6d}, the corresponding rotation matrix is constructed by
\begin{equation}
\mathbf{R}_i^{(l)} = \mathrm{rotmat}({\theta}_i^{(l)}), 
\quad
\mathbf{R}_i^{(l)} \in \mathrm{SO}(3),
\end{equation}
with the inverse mapping
\begin{equation}
{\theta}_i^{(l)} = \mathrm{rot6d}(\mathbf{R}_i^{(l)}),
\quad
{\theta}_i^{(l)} \in \mathbb{R}^6.
\end{equation}
Since relative knee rotations provide no useful information for ankle pose refinement, we transform them into global rotations in camera space.
For a joint $i$, this is done by multiplying the rotations of all its ancestor joints $\mathcal{A}(i)$ along the kinematic chain, from the root to the target joint:
\begin{equation}
\bar{\mathbf{R}}_i^{(l)} = \Big(\prod_{j \in \mathcal{A}(i)}^{\rightarrow} \mathbf{R}_j^{(l)}\Big) \mathbf{R}_i^{(l)}.
\end{equation}
The operator $\overset{\rightarrow}{\prod}$ denotes the ordered matrix multiplication along the chain, and $\bar{\theta}_i^{(l)} = \mathrm{rot6d}(\bar{\mathbf{R}}_i^{(l)})$ is the resulting global rotation of joint $i$ in 6D representation.
Conditioning on global knee rotations is crucial, as they constrain feasible ankle rotations and help disambiguate lifting 2D keypoint sequences to 3D rotations.
Initial ankle predictions act as a strong prior, and are especially important when 2D keypoints are noisy or missing.
Using global rather than knee-relative ankle rotations leads to FootMR better generalizing to extreme ankle motions.

\vspace{2mm}
\noindent\textbf{Network design.} Our network design is heavily inspired by GVHMR~\cite{shen2024gvhmr}.
The three per-frame input features $f_{\bar{\theta}_{k,a}}^{(l)}, f_{\text{foot2d}}^{(l)}, f_{\text{bbox}}^{(l)}$ are first independently mapped to the same dimension using dedicated MLPs.
They are then combined through element-wise summation to create one unified token $T^{(l)}\in\mathbb{R}^{d_\text{H}}$ per frame.
Subsequently, the resulting sequence is processed by multiple transformer encoder layers leveraging Rotary Position Embedding (RoPE)~\cite{su2024rope}, which encodes relative positional dependencies instead of absolute positions.
Relative positional embeddings together with a further introduced attention mask enable FootMR to process sequences of arbitrary length in a single forward pass, without relying on autoregressive inference strategies such as sliding-window.
The attention mask is constructed such that each token only attends to tokens within a $W$-frame neighborhood.
Due to the design decision to have only one token per frame, FootMR is computationally efficient even for larger $W$ and thus can effectively capture long-range dependencies.
The output tokens \mbox{$\{\tilde{T}^{(l)} \in \mathbb{R}^{d_\text{H}} \}^L_{l=1}$} of the final transformer encoder layer are independently processed by an MLP to predict the residual global rotation of the left and right ankle $\Delta\bar{\theta}_{\text{ankle}}^{(l)} \in \mathbb{R}^{2\times6}$.
Due to its efficient design and because only a few keypoints and rotations are processed per frame instead of heavy image features, our foot motion refinement is lightweight and introduces only minor computational overhead.

\vspace{2mm}
\noindent\textbf{Network output.}
The output of FootMR is simply added element-wise to the initial ankle predictions, producing the final global ankle rotations 
\begin{equation}
\bar{\theta}_{\text{ankle}}^{(l)} = \bar{\theta}_{\text{ankle}_\text{init}}^{(l)} + \Delta\bar{\theta}_{\text{ankle}}^{(l)}.
\end{equation}
To then convert the global to parent-relative ankle rotations, they must be multiplied by the inverse of the parent joint's rotation matrix:
\begin{equation}
\mathbf{R}_{\text{ankle}}^{(l)} = \left(\bar{\mathbf{R}}_{\text{knee}}^{(l)}\right)^{-1} \bar{\mathbf{R}}_{\text{ankle}}^{(l)}.
\end{equation}

\subsection{Training}
\label{sec:training}
We jointly train FootMR with the SMPL-X motion estimator from scratch.
We find that this produces slightly better results than training FootMR on top of a pretrained fixed-weight SMPL-X regressor.
We follow previous work~\cite{shin2024wham,shen2024gvhmr,shin2023bio} and synthesize input 2D keypoints during training by extracting 3D keypoints from the ground truth SMPL-X sequence, adding noise and projecting them onto a virtual camera.
Since image features are not used as input, additional data augmentation for training FootMR can be applied.
Specifically, for every sequence, we uniformly sample a random 3D rotation $\mathcal{R} \sim \mathcal{U}(\mathrm{SO}(3))$ and apply it to the ground truth and predicted root joint orientations.
This results in augmented foot keypoints, bounding boxes, and predicted global knee and ankle rotations.
By effectively simulating 3D foot motion in all possible orientations in 3D space, the variety of keypoints and rotations FootMR processes during training is heavily expanded.
This helps the model to generalize well to diverse foot motions.
Note that the random rotation data augmentation is only applied to the inputs of FootMR. 
The output does not need to be adjusted, as the updated global ankle rotations are always converted to parent-relative rotations that are independent of the orientation of the root.

We do not use any additional losses for training FootMR, but instead simply integrate its output into the losses of the SMPL-X motion estimation method.
Typical motion reconstruction losses influenced by the ankle rotations are
\begin{align}
\mathcal{L}_{\mathrm{j3d}} &= ||\mathcal{J}_{\mathrm{3d}}^{(l)} - {\hat{\mathcal{J}}}_{\mathrm{3d}}^{(l)}||_2 \\
\mathcal{L}_{\mathrm{j2d}} &= ||\Pi(\mathcal{J}_{\mathrm{3d}}^{(l)}) - {\hat{\mathcal{J}}}_{\mathrm{2d}}^{(l)}||_2 \\
\mathcal{L}_{\mathrm{v3d}} &= ||{V}_{\mathrm{3d}}^{(l)} - {\hat{V}}_{\mathrm{3d}}^{(l)}||_2 \\
\mathcal{L}_{\mathrm{v2d}} &= ||\Pi(V_{\mathrm{3d}}^{(l)}) - {\hat{V}}_{\mathrm{2d}}^{(l)}||_2 \\
\mathcal{L}_{\theta} &= \frac{1}{2} \left(||\theta^{*(l)} - \hat{\theta}^{(l)}||_2 + ||\theta^{(l)} - \hat{\theta}^{(l)}||_2\right)
\end{align}
which are calculated for every input frame $l$.
Locations of 3D joints and vertices of the SMPL-X model are denoted by $\mathcal{J}_{\mathrm{3d}}$ and ${V}_{\mathrm{3d}}$. The hat operator denotes the ground truth and $\Pi$ is the camera projection operator.
We only slightly adjust the joint rotation loss $\mathcal{L}_{\theta}$ such that it is calculated for the initial $\theta^{*(l)}$ and refined joint rotations $\theta^{(l)}$, which differ only in the ankle rotations.

\noindent\textbf{Architectural details.}
FootMR consists of six RoPE-based transformer encoder layers.
Each attention unit features four attention heads, and the hidden dimension is set to $d_\text{H} = 256$.
MLPs have two linear layers with GELU activation.
The maximum window size for self-attention is $W=120$.
This corresponds to a temporal window of four seconds when using videos with 30 frames per second.

%-------------------------------------------------------------------------------------------------------
\section{Experiments}
\label{sec:experiments}
%\vspace{2mm}
\noindent\textbf{Evaluation datasets.}
To quantitatively evaluate our approach, accurate 3D foot annotations are required.
Unfortunately, SMPL annotations provided by popular in-the-wild benchmark datasets such as 3DPW~\cite{marcard2018eccv} are often not accurate for the feet (see Fig.~\ref{fig:bad_gt_feet}, \textit{right}).
Thus, we instead evaluate on the multi-view datasets MOYO~\cite{tripathi2023moyo} and RICH~\cite{huang2022rich}.
MOYO~\cite{tripathi2023moyo} contains videos of a Yoga professional performing highly complex poses including extreme foot poses.
The subject is captured using a marker-based motion capture system, resulting in highly accurate ground truth SMPL-X fits.
RICH~\cite{huang2022rich} is recorded in both indoor and outdoor environments and reconstructs 3D human ground truth bodies using markerless motion capture. 
Multiple subjects are recorded performing a mixture of daily (\eg, cooking, cleaning, eating) and sporting activities (\eg, lunges, push-ups, burpees).
RICH contains significantly less foot motion than MOYO, and due to using markerless motion capture, the annotations are less accurate.
MOYO contains only a single subject and all videos are recorded in the same indoor environment.
Furthermore, the extreme out-of-domain body poses could lead to degenerated pose predictions, making it difficult to evaluate foot movement in isolation.
Due to these limitations we introduce the \textbf{MOOF} dataset (complex \textbf{MO}vements \textbf{O}f the \textbf{F}eet).
We record subjects performing simple body movements that involve complex foot motions, such as ankle circles, ankle stretches, and heel–toe walking, and augment these recordings with in-the-wild dance and ballet videos collected online.
MOOF comprises 15 subjects (9 female, 6 male) and 41 videos captured at 30 fps.
Video durations range from 4 seconds to 37~seconds, totaling 14,589 frames.
We use a semi-automatic annotation pipeline to obtain 2D ground truth keypoints for the big toe, small toe and heel.
Two example images from MOOF are shown in Fig.~\ref{fig:comparison}.
More details are provided in the supplementary material.

\vspace{2mm}
\noindent\textbf{Evaluation metrics.}
The accuracy of 3D human motion estimation is typically evaluated by computing the Mean Per Joint Position Error (MPJPE), Procrustes-aligned MPJPE (PA-MPJPE), and the Per Vertex Error (PVE).
Because joints or vertices of the feet represent only a small percentage of the entire body, improved foot poses have very little impact on the whole-body metrics.
Therefore, we instead compute the following foot-specific 3D metrics: the global Ankle Joint Angle Error (AJAE, in degrees) and the scale-normalized~\cite{rhodin2018cvpr} MPJPE for foot keypoints (big toe, small toe, heel), which we refer to as N-MPJPE\textsubscript{F} (reported in $\mathrm{mm}$).
Foot keypoints are mean-centered per foot before computing the N-MPJPE\textsubscript{F}.
Evaluation of 3D metrics is done in camera coordinates.
On the 2D dataset MOOF, we evaluate the 2D image alignment of reprojected foot keypoints using PCK\textsubscript{F} (Percentage of Correct foot Keypoints) and N-FKE\textsubscript{2d} (Normalized 2D Foot Keypoint Error).
PCK\textsubscript{F} measures the percentage of predicted foot keypoints with an L2 distance from the ground truth below a specified threshold.
We use a threshold of $0.05$, corresponding to $5\%$ of the person's bounding box.
Because PCK\textsubscript{F} also depends on the predicted shape, translation, and body pose, we additionally measure the N-FKE\textsubscript{2d}.
N-FKE\textsubscript{2d} is computed by first normalizing the keypoints using the person's bounding box, then mean-centering and scale-aligning them per foot, and finally computing the L2 distance from the ground truth.

\subsection{Implementation Details}
\noindent\textbf{SMPL-X motion estimator.}
To evaluate our foot motion refinement method, we couple it with the state-of-the-art 3D human motion recovery method GVHMR~\cite{shen2024gvhmr}.
We use GVHMR because it is efficient and jointly trainable with motion capture and video datasets.
GVHMR is a temporal transformer-based model that processes several input features including image features and 2D human keypoints.
We additionally create a baseline by slightly modifying GVHMR such that three 2D keypoints per foot (big toe, small toe, heel) are appended to the original 17 input body keypoints, and refer to the baseline as GVHMR\textsubscript{23j}.

\vspace{2mm}
\noindent\textbf{Training.}
We follow the training setting of GVHMR~\cite{shen2024gvhmr} and use the large-scale motion capture dataset AMASS~\cite{mahmood2019amass} and the video datasets BEDLAM~\cite{black2023bedlam}, Human3.6M~\cite{ionescu2014h36m}, and 3DPW~\cite{marcard2018eccv} for training.
We train FootMR together with GVHMR from scratch for 500 epochs with a batch size of 256, which takes around 30 hours on a single H100 GPU.
The AdamW optimizer~\cite{loshchilov2019adamw} is used with a learning rate of $2 \times 10^{-4}$ that is halved after 200 and 350 epochs.
Following \cite{shen2024gvhmr}, the losses presented in Sec.~\ref{sec:training} are weighted with $\lambda_{\theta} = 1$, $\lambda_{\mathrm{j3d}} = 500$, $\lambda_{\mathrm{j2d}} = 1000$, $\lambda_{\mathrm{v3d}} = 500$, $\lambda_{\mathrm{v2d}} = 1000$, and the length of the training sequences is set to $L=120$.
During training, ground truth global knee rotations are used as input for FootMR, as this slightly improves the convergence rate.
Predicted initial global ankle rotations are detached from the computational graph prior to being used as input.
We train a second model using the same setting by replacing GVHMR with GVHMR\textsubscript{23j}.

\begin{table*}% [t]\scriptsize
\newcolumntype{g}{>{\color{gray!100}}c}
	\centering
 \resizebox{1.0\textwidth}{!}{
	\begin{tabular}{clccg|ccg|cc}
		\toprule
        & & & MOYO & & & RICH & & \multicolumn{2}{c}{MOOF}\\
		\cmidrule(lr){3-5}\cmidrule(lr){6-8}\cmidrule(lr){9-10}
		  & Models &
		\multicolumn{1}{c}{AJAE$\downarrow$} & \multicolumn{1}{c}{N-MPJPE\textsubscript{F}$\downarrow$} & PA-MPJPE$\downarrow$ & \multicolumn{1}{c}{AJAE$\downarrow$} & \multicolumn{1}{c}{N-MPJPE\textsubscript{F}$\downarrow$} & PA-MPJPE$\downarrow$ & PCK\textsubscript{F} 0.05$\uparrow$ & N-FKE\textsubscript{2d}$\downarrow$  \\
		\midrule
        \parbox[t]{4mm}{\multirow{4}{*}{\rotatebox[origin=c]{90}{per-frame}}} & HMR2.0~\cite{goel2023humans} ICCV'23  & 42.0 & 55.7 & 83.5 & - & 36.9 & 57.2 & 76.1 & 1.95\\
        & ReFit~\cite{wang2023refit} ICCV'23 & 41.8 & 47.6 & 79.8 & - & 30.3 & 46.7 & 81.8 & 1.68\\
        & TokenHMR~\cite{dwivedi2024tokenhmr} CVPR'24  & 38.0 & 48.1 & 56.6 & - & 27.7 & 38.7 & 74.5 & 1.85\\
        & CameraHMR~\cite{patel2024camerahmr} 3DV'25  & 31.0 & 39.1 & 45.3 & - & 24.8 & 35.0 & 88.3 & 1.42\\
        \hline
        \parbox[t]{4mm}{\multirow{6}{*}{\rotatebox[origin=c]{90}{temporal}}} & WHAM~\cite{shin2024wham} CVPR'24  & 41.7 & 49.1 & 68.8 & - & 28.3 & 43.7 & 68.1 & 1.94 \\
        & TRAM~\cite{wang2024tram} ECCV'24  & 39.6  & 47.3 & 63.2 & - & 26.5 & 40.5 & 68.0 & 1.60 \\
        & GVHMR~\cite{shen2024gvhmr} SIGGRAPH'24  & 37.3 &  45.9 & 64.5 & 18.0 & 27.3 & 39.8 & 76.1 & 1.70\\
        \cmidrule{2-10}
        & GVHMR\textsubscript{23j} (Baseline) & 29.6 & 37.5 & 62.7 & 16.2 & 24.2 & 40.1 & \underline{89.2} & 1.06\\
        & GVHMR + \textbf{FootMR (Ours)}  & \underline{26.8} & \underline{32.8} & 62.2 & $\bm{15.8}$ & \underline{23.2} & 40.4 & 86.9 & $\bm{0.67}$\\
        & GVHMR\textsubscript{23j} + \textbf{FootMR (Ours)}  &  $\bm{25.9}$ & $\bm{32.4}$ & 62.3 & $\bm{15.8}$ & $\bm{23.0}$ &39.9 & $\bm{92.6}$ & $\bm{0.67}$\\
		\bottomrule
	\end{tabular}
    }
 	\caption{
    \textbf{Comparison of foot pose reconstruction} on the MOYO~\cite{tripathi2023moyo}, RICH~\cite{huang2022rich}, and MOOF datasets.
    Our foot motion refinement method (FootMR) effectively improves foot motion reconstruction of GVHMR and GVHMR\textsubscript{23j}, outperforming all competitors on all foot-specific metrics by a large margin.
    Note that improving full-body pose metrics is not the focus of this work; PA-MPJPE (shown in a faint column) is included only for reference.
    AJAE is in degrees, MPJPE in $\mathrm{mm}$, PCK\textsubscript{F} in $\mathrm{\%}$, and N-FKE in normalized pixel space.
  }
  	\label{table:main_results}
   %\vspace{-0.5em}
\end{table*}

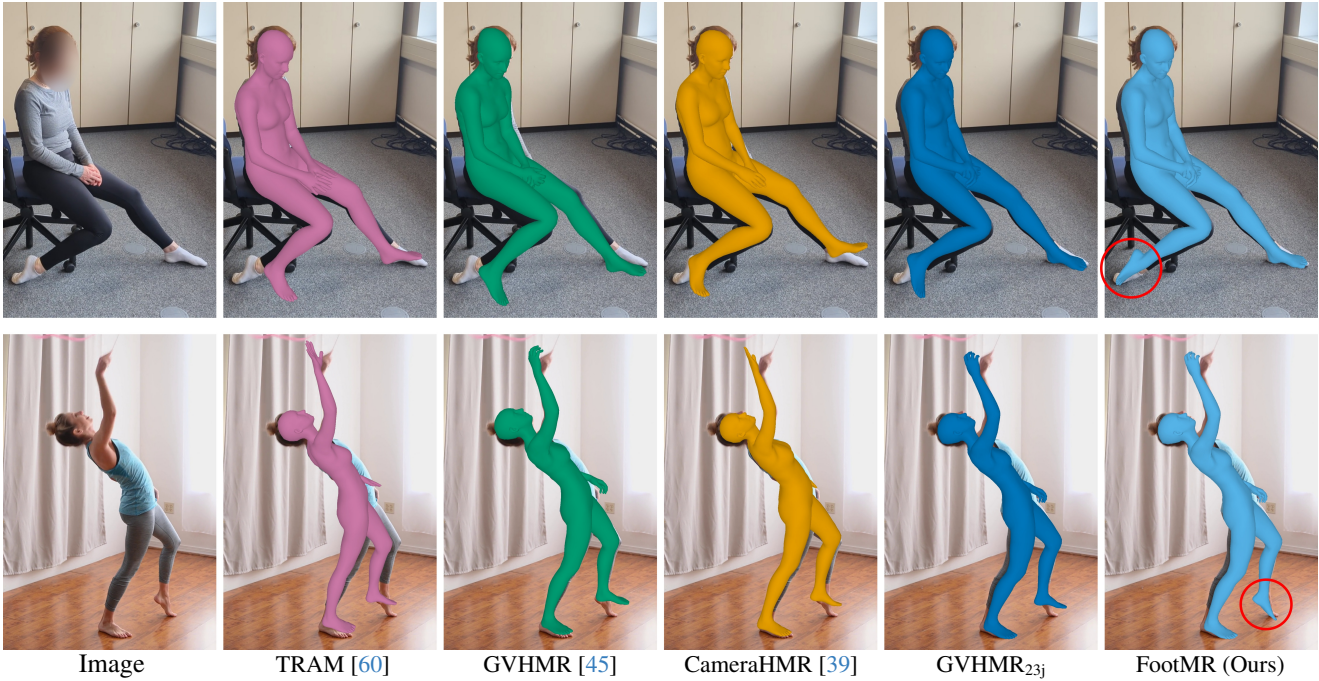
\begin{figure*}[t]
    \centering
    \begin{overpic}[width=\linewidth]{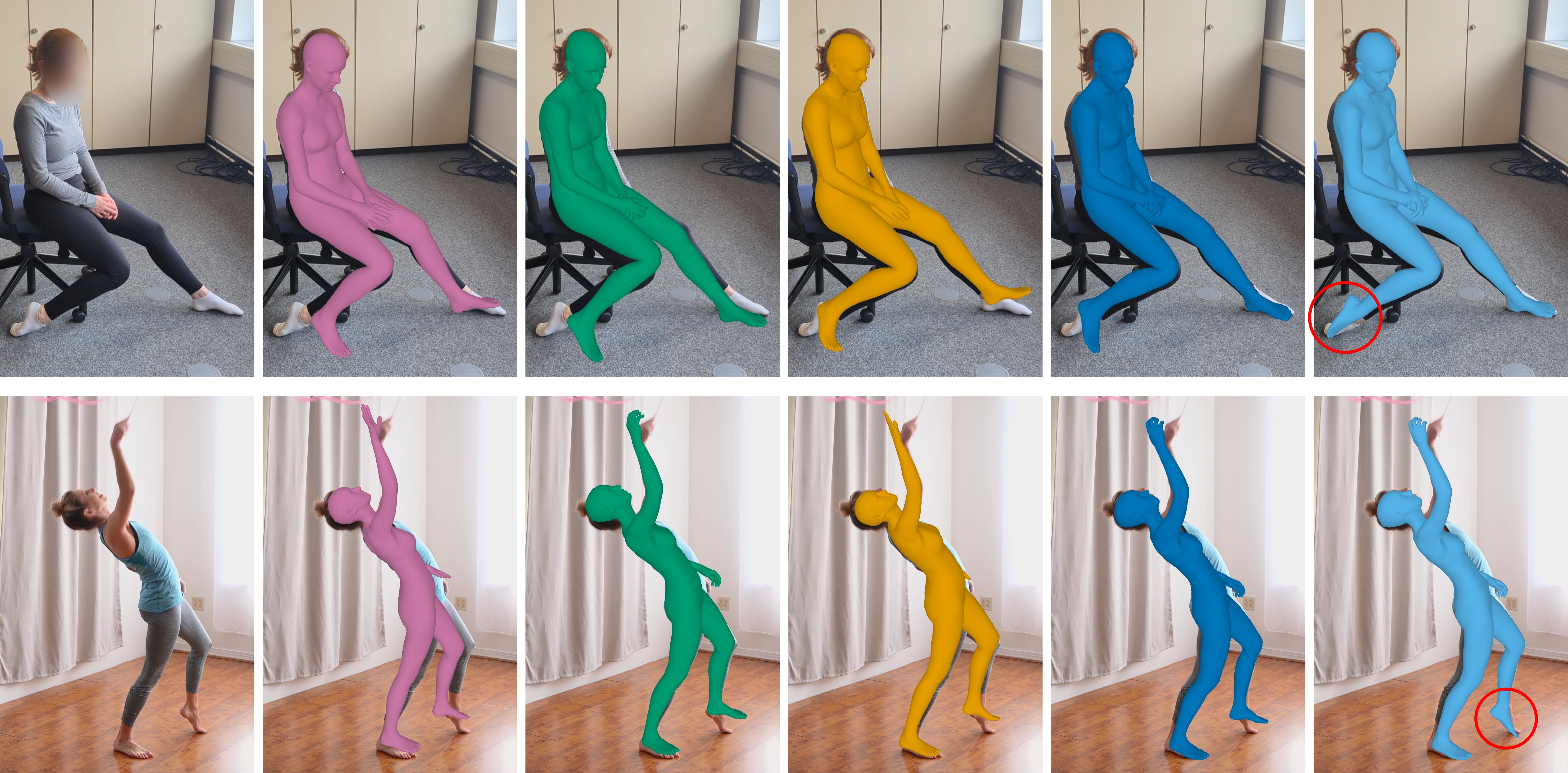}
        \put(5.7, -1.7){Image}
        \put(20.7, -1.7){\small TRAM~\cite{wang2024tram}}
        \put(36.5, -1.7){\small GVHMR~\cite{shen2024gvhmr}}
        \put(51.8, -1.7){\small CameraHMR~\cite{patel2024camerahmr}}
        \put(71, -1.7){\small GVHMR\textsubscript{23j}}
        \put(86.2, -1.7){\small FootMR (Ours)}
    \end{overpic}
    \vspace{0.1mm}
    \caption{
    \textbf{Qualitative comparison} on the introduced MOOF dataset. FootMR is the only method that is able to accurately reconstruct the extreme foot poses.
    }
    \label{fig:comparison}
    \vspace{-1.0em}
\end{figure*}

\subsection{Quantitative Evaluation}
We compare our approach with several state-of-the-art per-frame and video-based methods and with our straightforward baseline GVHMR\textsubscript{23j}.
For all competitors, we use the officially released model checkpoint to ensure a consistent and fair comparison.
No test-time flip augmentation is performed and ground truth focal lengths are not provided during inference.
Quantitative results on MOYO, RICH and MOOF are shown in Table~\ref{table:main_results}.
Our foot motion refinement method (FootMR) significantly improves foot motion reconstruction of both GVHMR and of our baseline GVHMR\textsubscript{23j}, outperforming all competitors by a large margin.
Notably, compared to the best existing temporal methods, FootMR achieves an error reduction of $30.6\%$ on MOYO ($37.3$ to $25.9$ AJAE) and of $58.1\%$ on MOOF ($1.60$ to $0.67$ N-FKE\textsubscript{2d}).
The same metrics are improved by $16.5\%$ and $52.8\%$ with respect to the best per-frame method~\cite{patel2024camerahmr}.
The strong performance on MOYO and MOOF shows that FootMR successfully generalizes to complex foot movements.
The reconstruction of everyday activities are also consistently improved as demonstrated on RICH.
Note that CameraHMR~\cite{patel2024camerahmr} achieves better full pose reconstructions (PA-MPJPE), primarily due to training on more diverse in-the-wild data with more accurate pseudo-GT annotations.
However, its foot metrics, especially on MOYO and MOOF, are significantly worse, showing that the improved pseudo-GT annotations are still not accurate and diverse enough for the feet.
The full-body pose metric PA-MPJPE is only shown for reference, as the focus of this work is solely on improving foot pose reconstruction.
Note that the ankle joint angle error (AJAE) cannot be evaluated on RICH for methods that predict SMPL parameters, since RICH only has ground truth in SMPL-X format.
All other metrics are computed by first converting the SMPL-X meshes to SMPL format using a vertex mapper~\cite{black2023bedlam}.

\begin{table}% [t]\scriptsize
	\centering
 %\setlength{\tabcolsep}{4pt}
 %\resizebox{0.87\linewidth}{!}{
 {\small{
	\begin{tabular}{lccc}
		\toprule
        & MOYO & RICH & MOOF \\
		\cmidrule{2-4}
		  Models &
		AJAE & AJAE & N-FKE\textsubscript{2d} \\
		\midrule
        Relative & 27.6 & 16.4 & 0.78 \\
        Global & 26.4 & 19.5 & 0.73\\
        Residual Relative & 28.7 & $\bm{15.8}$ & 0.90\\
        Residual Global (Ours) & $\bm{25.9}$ & $\bm{15.8}$ & $\bm{0.67}$\\
        \hline
        Ours w/o DA & 27.3 & 16.5 & 0.82 \\
		\bottomrule
	\end{tabular}
    }}
 	\caption{\textbf{Ablation of output rotation representation.}
    Estimating residual rotations with respect to initial global ankle rotations from a SMPL-X motion estimator leads to the best results.
  }
  	\label{table:ablation_outputs}
   \vspace{-0.5em}
\end{table}

\begin{table}[ht!]
\centering
\setlength{\tabcolsep}{3pt}
\renewcommand{\arraystretch}{1.1}
%\resizebox{0.48\textwidth}{!}
{\small{
\begin{tabular}{cccc|ccc}
\toprule
\multicolumn{4}{c}{Input Joint Rotations} & MOYO & RICH & MOOF \\
\cmidrule{1-7}
\multicolumn{1}{c}{Ankle} & \multicolumn{1}{c}{Knee} & \multicolumn{1}{c}{Hip} & \multicolumn{1}{c}{Pelvis} & \multicolumn{1}{c}{AJAE} & \multicolumn{1}{c}{AJAE} & \multicolumn{1}{c}{N-FKE\textsubscript{2d}} \\
\hline % \cmidrule{1-7}

$\times$ & $\times$ &  $\times$ & $\times$ & 38.0 & 36.0 & 0.65 \\
  $\surd$ & $\times$ &$\times$ & $\times$ & 29.1 & 17.4 & $\bm{0.61}$  \\
$\surd$ & $\surd$ & $\times$ & $\times$ & $\bm{25.9}$ & 15.8 & 0.67 \\
$\surd$  & $\surd$ & $\surd$ & $\times$ & 26.1 & $\bm{15.7}$ & 0.71 \\
$\surd$ & $\surd$ & $\surd$ &  $\surd$ & 26.0 & $\bm{15.7}$ & 0.71  \\

\bottomrule
\end{tabular}
}}
\caption{\textbf{Ablation of input joint rotations.}
Using global knee and ankle rotations as input for foot motion refinement is important, while additional joints provide no further benefit.
}
\label{table:ablation_inputs}
\vspace{-1em}
\end{table}

Interestingly, our simple baseline GVHMR\textsubscript{23j} already outperforms all competitors on all foot metrics.
Intuitively, by additionally providing the three 2D keypoints per foot, GVHMR\textsubscript{23j} can fully exploit large-scale motion capture data during training to learn about 3D foot motion.
However, unlike FootMR, it still cannot generalize to extreme foot poses, which is evident in the poorer performance on MOOF (N-FKE\textsubscript{2d} of $1.06$ vs.\ $0.67$) and also visually shown in Fig.~\ref{fig:comparison}.
Combining FootHMR with GVHMR\textsubscript{23j} achieves slightly better results than combining it with GVHMR.
We notice that using the additional 2D foot keypoints leads to GVHMR\textsubscript{23j} estimating more accurate knee joint rotations.
For instance, the knee joint angle error on MOYO improves from $20.6^\circ$ to $18.9^\circ$.
This facilitates the ankle joint refinement task and results in better 2D image alignment, as can be seen by the PCK\textsubscript{F}.

Qualitative comparisons with state-of-the-art competitors are shown in Fig.~\ref{fig:comparison}.
All methods predict similar body poses, but only FootMR is able to accurately reconstruct the extreme foot poses.
For more qualitative results, please refer to our supplementary materials.

\subsection{Ablation Study}
To analyze the impact of each component of our method, we evaluate multiple variants of FootMR combined with GVHMR\textsubscript{23j} using the same training and evaluation protocol.
We begin by investigating the influence of different rotation representations of the ankle rotations predicted by FootMR.
The results are shown in Table~\ref{table:ablation_outputs}.
The models in the first two rows receive global knee rotations as input and directly estimate the parent-relative or global ankle rotations, instead of residual rotations.
When estimating parent-relative rotations, the model cannot generalize well to extreme foot poses, which is reflected in the poorer performance on MOYO and MOOF.
The reason could be the limited range of ankle rotations observed during training.
Directly estimating global ankle rotations leads to performance degradation on RICH, especially when 2D keypoints are missing due to occlusions.
Combining the best of both worlds, FootMR estimates residual rotations with respect to initial global ankle rotations from a SMPL-X motion estimator.
This significantly improves robustness to noisy or missing 2D keypoints, while generalizing well to extreme foot poses.
It also achieves better results than predicting residual rotations with respect to initial parent-relative ankle rotations.
Table~\ref{table:ablation_outputs} additionally shows the impact of the proposed random rotation data augmentation technique, which consistently improves all foot metrics.

Next, we analyze which joint rotations are important as input for refining the ankle rotations.
The ablation study is shown in Table~\ref{table:ablation_inputs}.
A model that directly predicts global ankle rotations only from 2D foot keypoint sequences heavily degenerates when input keypoints are noisy, leading to very poor results on MOYO and RICH.
This demonstrates the high ambiguity of the 2D-to-3D mapping.
Interestingly, it still manages to achieve good 2D image alignment, showing that 2D foot metrics alone are no reliable indicator to assess the 3D reconstruction performance.
Using initial global ankle rotations as input for refinement leads to less ambiguity.
Global knee rotations further help to disambiguate the 2D keypoint sequences to 3D rotations mapping.
Finally, providing additional joints along the kinematic chain (hip and pelvis) offers no further advantages.
More ablation studies are included in the supplementary material.

\section{Limitations}
FootMR achieves state-of-the-art foot motion reconstruction by accurately capturing the movement of the left and right ankle joint.
However, a single ankle joint is not sufficient to model the full range of motion of the foot.
Additionally improving the forefoot joint of SMPL-X only helps to a limited extent.
Feet on the SMPL-X body are overly simplistic and cannot model movements such as curling of the toes \cite{osman2022supr}.
Future work should explore extending our approach to the articulated foot model SUPR-Foot~\cite{osman2022supr} which contains 13 joints per foot.
More dense 2D keypoints will be needed to estimate the additional degrees of freedom.

%-------------------------------------------------------------------------------------------------------
\section{Conclusion}
\label{sec:conclusion}
FootMR is a novel approach for improving 3D foot motion reconstruction in markerless monocular human motion capture.
It refines initial ankle rotation estimates from an existing 3D human recovery model by leveraging detected 2D foot keypoints.
Our experiments show that incorporating knee and initial ankle rotations and predicting only residual ankle rotations effectively resolves ambiguity of lifting the 2D foot keypoint sequences to 3D.
By not using images directly as input, FootMR does not suffer from inaccurate foot annotations present in most existing human datasets.
It also enables our proposed random root orientation data augmentation which further improves generalization to extreme foot poses.
To support evaluation, we introduced MOOF, a new video dataset containing complex foot motion.
FootMR combined with the 3D human motion recovery model GVHMR produces accurate and temporally coherent 3D human body and foot motion, outperforming all competitors across diverse benchmarks on various foot-specific metrics.
Most notably, FootMR is the only method that is able to accurately reconstruct extreme foot poses not seen in common datasets.

\vspace{-1.0em}
\small{\paragraph{Acknowledgements.}
This work was supported by the Federal Ministry of Education and Research (BMBF), Germany, under the AI service center KISSKI (grant no.\ 01IS22093C), the Deutsche Forschungsgemeinschaft (DFG) under Germany’s Excellence Strategy within the Cluster of Excellence PhoenixD (EXC 2122), and the European Union within the Horizon Europe research and innovation programme under grant agreement no.\ 101136006 – XTREME.
}

{
    \small
    \bibliographystyle{ieeenat_fullname}
    \bibliography{main}
}

% --- Start of Supplementary Material ---
\cleardoublepage
\appendix

% 1. FIX FOR "COMMAND UNDEFINED" AND JUMPING LINKS
% This creates a unique "base" for all hyperref anchors in the appendix
\makeatletter
\newcommand{\hb@xt@Patch}{%
  \renewcommand{\theHsection}{supp.\thesection}%
  \renewcommand{\theHfigure}{supp.\thefigure}%
  \renewcommand{\theHtable}{supp.\thetable}%
  \renewcommand{\theHequation}{supp.\theequation}%
}
\hb@xt@Patch
\makeatother

% 2. Reset counters
\setcounter{section}{0}
\setcounter{subsection}{0}
\setcounter{figure}{0}
\setcounter{table}{0}
\setcounter{equation}{0}

% 3. Redefine visual numbering formats
\renewcommand\thesection{\Alph{section}}
\renewcommand\thesubsection{\thesection.\arabic{subsection}}
\renewcommand{\thefigure}{S\arabic{figure}}
\renewcommand{\thetable}{S\arabic{table}}
\renewcommand{\theequation}{S\arabic{equation}}

%----------------------------------------------------------------------------------------------------------------------
\section{MOOF Dataset}
To support the evaluation of foot motion reconstruction, we propose a novel dataset called MOOF, short for complex \textbf{MO}vements \textbf{O}f the \textbf{F}eet.
We recorded 3 subjects (1 female, 2 male) in an indoor environment performing simple body movements that involve complex foot motions.
The recordings were approved by our institution’s ethics committee.
All participants volunteered and provided written consent for the capture, use, and distribution of their data for research purposes.
The recordings were de-identified by obscuring the faces of all subjects to ensure privacy protection.
A total of 22 videos were recorded with a static smartphone camera at a resolution of $1920\times1080$~pixels and 30 frames per second.
In addition, we collected dance and ballet videos with a focus on diverse foot movements from stock footage providers~\cite{pexels}.
The videos contain different indoor and outdoor scenes and were recorded with static or dynamic cameras.
In total, MOOF comprises 15~subjects (9~female, 6~male) and 41~videos captured at 30~fps.
A detailed description of the sequences that appear in MOOF is shown in Table~\ref{table:moof}.

\begin{figure*}
\centering
\includegraphics[width=1.0\linewidth]{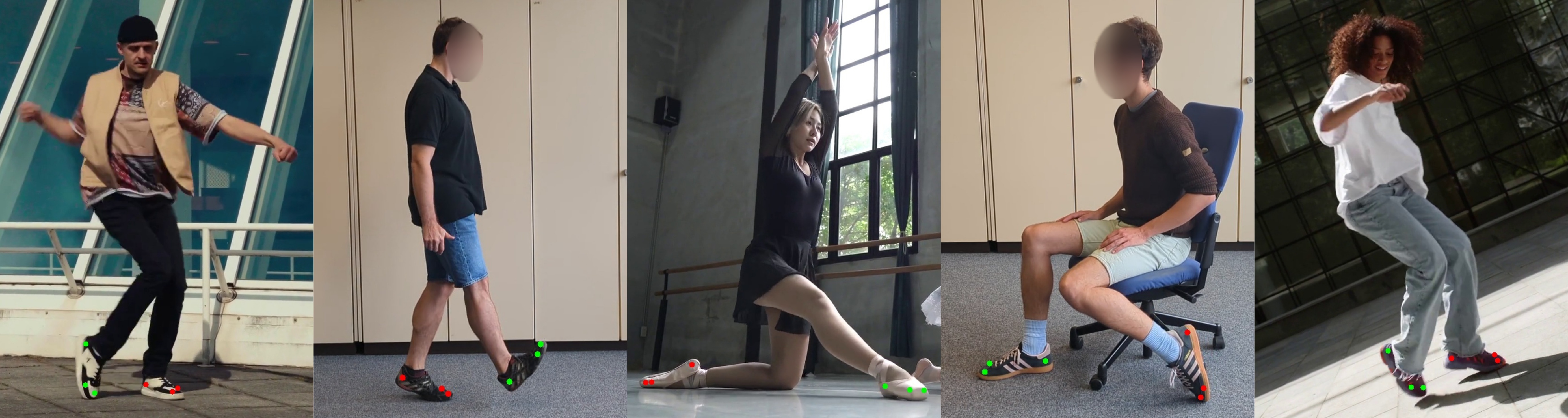}
\caption{\textbf{Video frames from MOOF.} The MOOF dataset consists of videos with annotated 2D keypoints for the left and right big toe, small toe, and heel.
The collected videos are designed to highlight complex foot movements, making the dataset well-suited for evaluating fine-grained foot motion reconstruction in scenarios such as dance, sports, and rehabilitation.
}
\label{fig:moof}
%\vspace{-1.0em}
\end{figure*}

We label three keypoints per foot: big toe, small toe, and heel.
Following \cite{jin2020whole}, keypoints are defined in the inner center rather than on the surface.
To speed up the annotation process, we use a semi-automatic annotation pipeline.
First, YOLOv8~\cite{jocher2023yolo} tracks the bounding box of the target person in each video.
The corresponding image crops are then used as input for the 2D pose estimator Sapiens-2B~\cite{khirodkar2024sapiens}, which detects 133 keypoints, including the foot keypoints, following COCO-WholeBody~\cite{jin2020whole}.
We manually inspect all foot keypoint detections, correcting faulty ones and adding a binary visibility label.
Because we recorded and selected videos where feet are clearly visible in almost all frames, and due to the high detection accuracy of Sapiens-2B, only 2,898~keypoints ($3.3\%$) needed to be manually corrected.
In summary, MOOF provides 14,589 video frames, each annotated with the person’s bounding box as well as 2D keypoint and visibility labels for the left and right big toe, small toe, and heel.
Example video frames are shown in Fig.~\ref{fig:moof}.

\section{Implementation Details}
\label{sec:impl_suppl}
%\vspace{2mm}
\noindent\textbf{Experiments.}
We use the state-of-the-art 2D whole-body pose estimator Sapiens-2B~\cite{khirodkar2024sapiens} to estimate the 2D foot keypoints during inference.
The 17 body joints used by GVHMR~\cite{shen2024gvhmr} are also included in the output of Sapiens.
However, to ensure a consistent comparison with GVHMR, we follow their setting and use ViTPose~\cite{xu22vitpose} to detect the body keypoints.
The introduced baseline GVHMR\textsubscript{23j} appends the six foot keypoints to the 17 body keypoints and is then trained following the exact training setting of GVHMR. 

\begin{table}% [t]\scriptsize
	\centering
 \resizebox{0.85\linewidth}{!}{
	\begin{tabular}{cccc}
		\toprule
           Activity & \# Subjects  &   \# Videos & \# Frames \\
		\midrule
        ankle circles & 3 & 6 & 1,640 \\
        ankle stretches & 3 & 6& 1,399 \\
        ballet & 2 & 6 & 1,808 \\
        dance & 9 & 12 & 6,509 \\
        gymnastics & 2 & 2 & 783 \\
        tapping feet & 3 & 4 & 1,186 \\
        heel-toe walk & 3 & 5 & 1,264 \\
        \midrule
        \textbf{Total} & 15 & 41 & 14,589 \\
		\bottomrule
	\end{tabular}
    }
 	\caption{\textbf{Detailed description of the MOOF dataset.} Some of the subjects perform multiple activities. The videos are recorded at 30 fps.
  }
  	\label{table:moof}
   %\vspace{-0.5em}
\end{table}

\vspace{2mm}
\noindent\textbf{Runtime.} We evaluate the runtime on an example video of 1,123 frames (37.4 seconds) using an RTX 4090 GPU.
Without feature preprocessing, FootMR combined with GVHMR~\cite{shen2024gvhmr} takes 0.192 seconds to process the video, compared to 0.182 seconds when only GVHMR is used.
Thus, FootMR adds merely 10 milliseconds of overhead when processing the full sequence of 1,123 frames.
This is orders of magnitude faster than optimization-based refinement methods such as SMPLify~\cite{bogo16eccv}, which can require up to several seconds per frame.
However, GVHMR and FootMR rely on the output of several other methods, including bounding box detection~\cite{jocher2023yolo}, image feature extraction~\cite{goel2023humans}, and 2D keypoint detection~\cite{xu22vitpose,khirodkar2024sapiens} (see runtime analysis reported by GVHMR~\cite{shen2024gvhmr}).
GVHMR uses ViTPose~\cite{xu22vitpose} to detect 17 body keypoints, which takes 19.3~seconds to process the video.
To estimate the 2D foot keypoints, we use the state-of-the-art 2D whole-body pose estimator Sapiens-2B~\cite{khirodkar2024sapiens}.
Due to processing images at a high resolution of $1024\times768$ pixels, running Sapiens on the video takes around 648 seconds.
If realtime performance is of interest, a promising research direction would be to train an efficient 2D foot keypoint detector via knowledge distillation~\cite{hinton2015distilling}.
By only focusing on foot keypoints, a smaller crop around the foot could be used as input instead of the high resolution whole-body crop, which would significantly reduce the runtime. 

Training GVHMR jointly with FootMR from scratch takes around 30 hours on a single H100 GPU.
In comparison, the training of GVHMR without FootMR is completed after approximately 27.5 hours, resulting in a training overhead of FootMR of 2.5 hours.

\section{Additional Results}
\noindent\textbf{Temporal smoothness.}
To evaluate the inter-frame smoothness of reconstructed foot motion, we compute the Acceleration error of the 3D foot keypoints (Accel\textsubscript{F}, in $\mathrm{m / s^2}$).
The results together with the per-frame accuracy metric N-MPJPE\textsubscript{F} are reported in Table~\ref{table:accel}.
All temporal methods produce significantly smoother 3D foot motion than the per-frame methods.
FootMR achieves the lowest acceleration and per-frame reconstruction error on RICH.
On MOYO, WHAM has a slightly lower acceleration error than FootMR.
However, its per-frame reconstruction error is significantly higher, indicating that WHAM potentially over-smoothes the foot motion.
Overall, FootMR produces more accurate foot motion than prior work, with a level of smoothness on par with existing temporal methods.

\begin{table}% [t]\scriptsize
	\centering
 \resizebox{\linewidth}{!}{
	\begin{tabular}{clcc|cc}
		\toprule
        & &  \multicolumn{2}{c}{MOYO} &  \multicolumn{2}{c}{RICH} \\
		\cmidrule(lr){3-4}\cmidrule(lr){5-6}
		  & Models &
		\multicolumn{1}{c}{Accel\textsubscript{F}} & \multicolumn{1}{c}{N-MPJPE\textsubscript{F}}  & \multicolumn{1}{c}{Accel\textsubscript{F}} & \multicolumn{1}{c}{N-MPJPE\textsubscript{F}} \\
		\midrule
        \parbox[t]{4mm}{\multirow{4}{*}{\rotatebox[origin=c]{90}{per-frame}}} & HMR2.0~\cite{goel2023humans} & 16.9 & 55.7 & 21.7 & 36.9 \\
        & ReFit~\cite{wang2023refit}  & 37.0 & 47.6 & 30.9 & 30.3\\
        & TokenHMR~\cite{dwivedi2024tokenhmr}   & 19.4 & 48.1 & 26.0 & 27.7\\
        & CameraHMR~\cite{patel2024camerahmr}  & 13.8 & \underline{39.1} & 22.9 &  \underline{24.8} \\
        \hline
        \parbox[t]{4mm}{\multirow{4}{*}{\rotatebox[origin=c]{90}{temporal}}} & WHAM~\cite{shin2024wham} & $\bm{2.5}$ & 49.1  & 8.1 & 28.3 \\
        & TRAM~\cite{wang2024tram}  & 4.3 & 47.3 & 9.3 &  26.5  \\
        & GVHMR~\cite{shen2024gvhmr}  &3.7 &  45.9 & \underline{7.5} & 27.3 \\
        & \textbf{FootMR (Ours)}  & \underline{3.6} & $\bm{32.4}$ & $\bm{7.3}$ &  $\bm{23.0}$ \\
		\bottomrule
	\end{tabular}
    }
 	\caption{
    \textbf{Comparison of temporal smoothness} of reconstructed foot motion on MOYO~\cite{tripathi2023moyo} and RICH~\cite{huang2022rich}.
    Accel\textsubscript{F} is in $\mathrm{m / s^2}$ and N-MPJPE\textsubscript{F} in $\mathrm{mm}$.
  }
  	\label{table:accel}
   %\vspace{-0.5em}
\end{table}

\vspace{2mm}
\noindent\textbf{Ablation study.} 
Further ablation studies are presented in Table~\ref{table:ablations_suppl}.
The first row shows the performance of our model when evaluated with 2D foot keypoints from ViTPose-WholeBody~\cite{xu2022vitpose+} instead of Sapiens-2B.
ViTPose-WholeBody is significantly faster with roughly the same runtime as ViTPose~\cite{xu22vitpose} (see runtime analysis in Sec.~\ref{sec:impl_suppl}).
Due to the less accurate 2D foot keypoints, the performance drops on MOYO and slightly decreases on RICH and MOOF.
This shows the importance of accurate 2D~foot~keypoints for 3D foot motion reconstruction.
Future work could investigate developing an efficient and accurate 2D~foot keypoint detector.
However, using the less accurate 2D foot keypoints from ViTPose-WholeBody still leads to significant improvements over GVHMR.
Next, we train FootMR\textsubscript{23j}, a version of FootMR that takes all body joints in addition to the foot keypoints as input.
Evaluation results show that the 2D body keypoints are not useful for foot motion refinement.
FootMR\textsubscript{23j} achieves slightly worse results on MOYO and MOOF, indicating that the additional body keypoints potentially hinder generalization to extreme foot poses.

As described in the main paper, a major challenge is that current training datasets contain inaccurate image-3D foot annotation pairs.
FootMR addresses this by refining foot motion decoupled from the rest of the body and only utilizing 2D keypoints instead of images.
As an alternative approach, we train a version of GVHMR\textsubscript{23j} that does not use image features as input and thus also completely bypasses the dependency on inaccurate foot annotations.
The results in Table~\ref{table:ablations_suppl} demonstrate that FootMR significantly outperforms this baseline in foot and body pose metrics.
It shows that image features are important for accurately reconstructing the body pose.
Furthermore, predicting ankle rotations decoupled from the remaining body joints helps to address the issue of spurious correlations, leading to a model that better generalizes to extreme foot poses.
Finally, we report the results for FootMR trained on top of a pretrained fixed-weight GVHMR\textsubscript{23j} model.
This produces slightly worse results than jointly training both models from scratch.

\begin{table*}% [t]\scriptsize
	\centering
 \resizebox{1.0\textwidth}{!}{
	\begin{tabular}{lccc|ccc|cc}
		\toprule
        & & MOYO & & & RICH & & \multicolumn{2}{c}{MOOF}\\
		\cmidrule(lr){2-4}\cmidrule(lr){5-7}\cmidrule(lr){8-9}
		  Models &
		\multicolumn{1}{c}{AJAE$\downarrow$} & \multicolumn{1}{c}{N-MPJPE\textsubscript{F}$\downarrow$} & \multicolumn{1}{c}{PA-MPJPE$\downarrow$} & \multicolumn{1}{c}{AJAE$\downarrow$} & \multicolumn{1}{c}{N-MPJPE\textsubscript{F}$\downarrow$} & \multicolumn{1}{c}{PA-MPJPE$\downarrow$} & PCK\textsubscript{F} 0.05$\uparrow$ & N-FKE\textsubscript{2d}$\downarrow$  \\
		\midrule
        ViTPose Foot Kpts &  {29.2} & {35.7} & 63.4 & {16.0} & {23.3} & 40.2 & {91.8} & {0.76}\\
        FootMR\textsubscript{23j} & \underline{27.0} & 33.6 & 62.2 & $\bm{15.7}$ & \underline{23.1} & 40.1 & \underline{92.5} & 0.71 \\
        \hline
        GVHMR\textsubscript{23j} w/o image feature & 30.8 & 39.2 & 86.4 & 18.1 & 26.4 & 50.2 & 92.4 & 0.95\\
        GVHMR\textsubscript{23j} frozen + FootMR & 27.5 & \underline{33.2} & 62.0 & 16.9 & 24.2 & 40.2 & 92.2 & \underline{0.70}\\
        \hline
        GVHMR~\cite{shen2024gvhmr} & 37.3 &  45.9 & 64.5 & 18.0 & 27.3 & 39.8 & 76.1 & 1.70\\
        GVHMR\textsubscript{23j} + FootMR \textbf{(Ours)}  &  $\bm{25.9}$ & $\bm{32.4}$ & 62.3 & \underline{15.8} & $\bm{23.0}$ &39.9 & $\bm{92.6}$ & $\bm{0.67}$ \\
		\bottomrule
	\end{tabular}
    }
 	\caption{
    \textbf{Additional ablation studies} on the MOYO~\cite{tripathi2023moyo}, RICH~\cite{huang2022rich}, and MOOF datasets.
    Please see the text for details.
    AJAE is in degrees, MPJPE in $\mathrm{mm}$, PCK\textsubscript{F} in $\mathrm{\%}$, and N-FKE in normalized pixel space.
  }
  	\label{table:ablations_suppl}
   %\vspace{-0.5em}
\end{table*}

\begin{figure*}[t]
    \centering
    \begin{overpic}[width=0.89\linewidth]{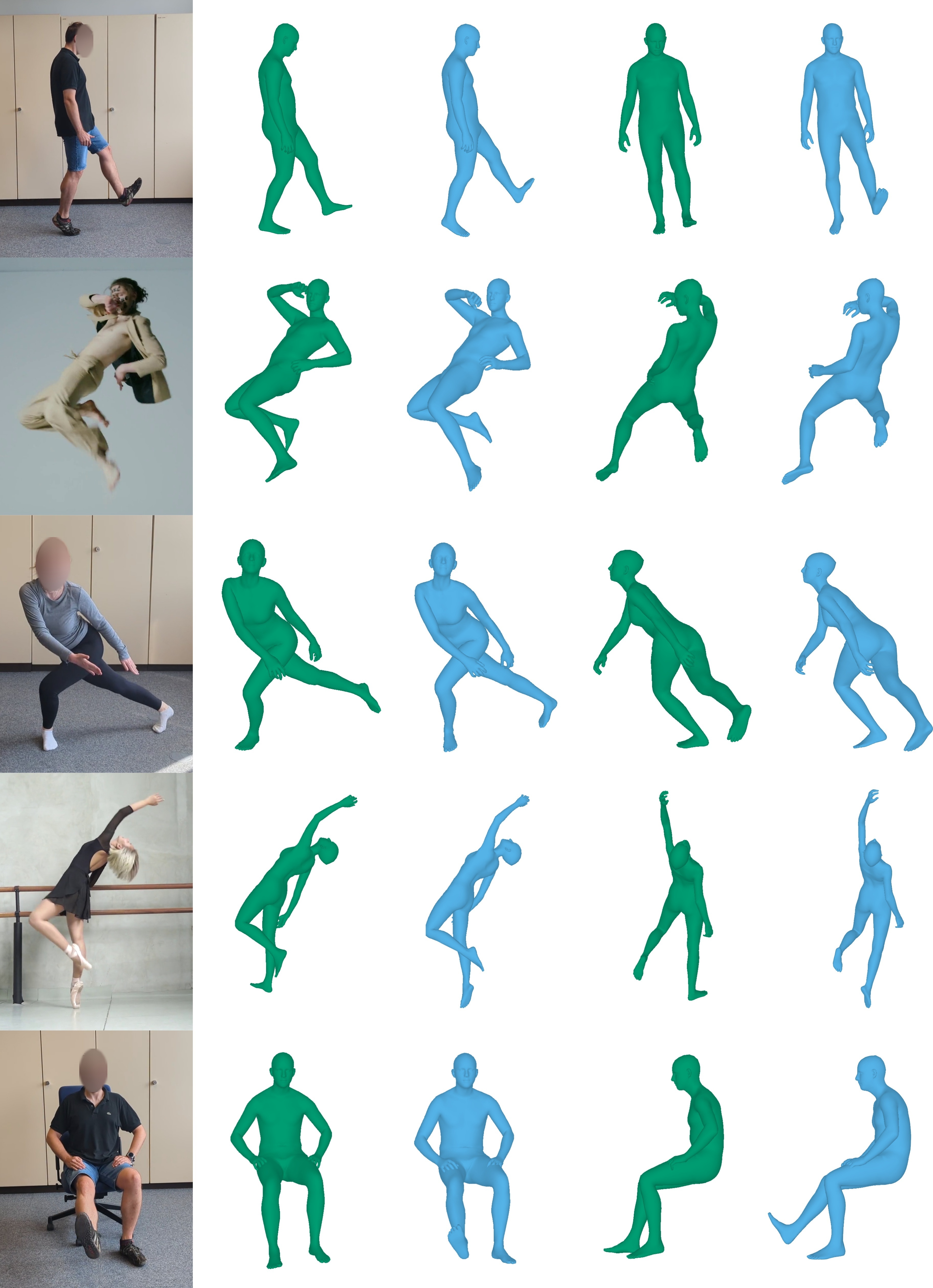}
        \put(5.5, -1.5){Image}
        \put(19.3, -1.5){\small GVHMR~\cite{shen2024gvhmr}}
        \put(33.1, -1.5){\small FootMR (Ours)}
        \put(52.0, -1.5){\small Alternative Viewpoint}
    \end{overpic}
    \vspace{3mm}
    \caption{
    \textbf{Qualitative comparison on MOOF.} \textcolor{sky_blue}{FootMR} is able to reconstruct more accurate foot poses than the competitor \textcolor{bluish_green}{GVHMR}~\cite{shen2024gvhmr}. 
    }
    \label{fig:qual_res}
    %\vspace{-1.0em}
\end{figure*}

\vspace{2mm}
\noindent\textbf{Qualitative results.} 
Additional qualitative comparisons are presented in Fig.~\ref{fig:qual_res}.
Since static images cannot convey the temporal accuracy and realism of the results, we also encourage the reader to view the supplementary video on our project page at \href{https://twehrbein.github.io/footmr-website/}{twehrbein.github.io/footmr-website/}.

\section{Future Work}
FootMR refines 3D foot motion in camera coordinates.
To recover global human motion, existing approaches transform camera-space motion to world-space utilizing learned motion priors and SLAM methods~\cite{sun2023trace,wang2024tram,yin2024whac,ye2023slahmr}, or by directly regressing per-frame global orientation and translation~\cite{shin2024wham,shen2024gvhmr}.
GVHMR~\cite{shen2024gvhmr} additionally predicts stationary labels for hands and feet, which are used to postprocess the global motion to refine foot sliding and global trajectories.
Although compatible with our approach, we find that the stationary joint predictor tends to classify fine-grained foot movements as static, resulting in such movements often being suppressed in postprocessing.
Future work could investigate improving the stationary label predictions by leveraging FootMR's foot motion estimates.
Note that all quantitative and qualitative results presented in this work are in camera-space without any postprocessing.

Finally, an interesting research direction could be to develop a version of FootMR that can be used off-the-shelf in combination with any 3D human mesh recovery model.
One could explore leveraging initial ankle rotation estimates from diverse human recovery methods during training or augmenting with synthetically perturbed ankle rotations.

\end{document}